\documentclass[a4paper,fleqn]{cas-dc}

\usepackage[numbers]{natbib}
\usepackage{epstopdf}
\usepackage{color}
\usepackage{varwidth}
\definecolor{mycolor}{rgb}{0.53,0.59,0.65}
\usepackage{subfigure}
\usepackage[linesnumbered,ruled,vlined]{algorithm2e}

\usepackage{caption}

\usepackage[
singlelinecheck=false % <-- important
]{caption}

\def\tsc#1{\csdef{#1}{\textsc{\lowercase{#1}}\xspace}}
\tsc{WGM}
\tsc{QE}
\tsc{EP}
\tsc{PMS}
\tsc{BEC}
\tsc{DE}
\begin{document}
\let\WriteBookmarks\relax
\def\floatpagepagefraction{1}
\def\textpagefraction{.001}
\shorttitle{}
\shortauthors{Y. Wu, Y. Wang, X. Huang, F. Yang, S Ling, S Su, et~al.} %% 缩略作者 自己名字， 比如： 张三 = S. Zhang

%% 标题
\title [mode = title]{Multimodal Learning for Non-small Cell Lung Cancer Prognosis}                      
%%\tnotemark[1,2]

%%\tnotetext[1]{This document is the results of the research project funded by the National Science Foundation.}

%%\tnotetext[2]{The second title footnote which is a longer text matter to fill through the whole text width and overflow into another line in the footnotes area of the first page.}

%% 作者顺序
%% 1
\author[1]{\textcolor[RGB]{0,0,1}{Yujiao Wu}}[orcid=0000-0001-6366-9834]
\ead{yujiaowu111@gmail.com}
%\fnmark[1] %%第几作者
%\credit{}%%本文的贡献

%% 2
\author[2]{\textcolor[RGB]{0,0,1}{Yaxiong Wang}}[orcid=0000-0001-6596-8117]
\ead{wangyx15@stu.xjtu.edu.cn}
%%3
\author[3]{\textcolor[RGB]{0,0,1}{Xiaoshui Huang}}[orcid=0000-0002-3579-538X]
\ead{huangxiaoshui@pjlab.org.cn}
%\fnmark[3]
%% 4
\author[4]{\textcolor[RGB]{0,0,1}{Fan Yang}}[orcid=0000-0003-4017-4577]
\ead{yangfan@wpeony.com}

%\fnmark[4]

\author[5]{\textcolor[RGB]{0,0,1}{Sai Ho Ling}}[orcid=0000-0003-0849-5098]
\ead{Steve.Ling@uts.edu.au}

\author[6]{\textcolor[RGB]{0,0,1}{Steven Weidong Su}}[orcid=0000-0002-5720-8852]
\ead{Steven.Su@uts.edu.au}
\cormark[1]
%\fnmark[5]
\cormark[1]%%通讯作者星标
\address[1]{Peng Cheng Laboratory, Shenzhen, China}
\address[2]{School of Software Engineering, Xi'an Jiaotong University, Xi'an, 710049, China}  
\address[3]{Shanghai AI Laboratory, Shanghai, China} 
\address[4]{Shenzhen Peini Digital Technology Co., Ltd.} 
\address[5]{Electrical and Data Engineering, University of Technology Sydney, 15 Broadway, Ultimo NSW 2007, Australia}
\address[6]{Biomedical Engineering, University of Technology Sydney, 15 Broadway, Ultimo NSW 2007, Australia}

\cortext[cor1]{Corresponding author} %% 首页左下角通讯作者

%%\cortext[cor2]{Principal corresponding author} 

%%\fntext[fn1]{This is the first author footnote. but is common to thirdauthor as well.}
%%\fntext[fn2]{Another author footnote, this is a very long footnote and it should be a really long footnote. But this footnote is not yet sufficiently long enough to make two lines of footnote text.}

%%\nonumnote{This note has no numbers. In this work we demonstrate $a_b$ the formation Y\_1 of a new type of polariton on the interface between a cuprous oxide slab and a polystyrene micro-sphere placed on the slab.}

%%摘要
\begin{abstract}
 This paper focuses on the task of survival time analysis for lung cancer. Although much progress has been made in this problem in recent years, the performance of existing methods is still far from satisfactory. Traditional and some deep learning- based survival time analyses for lung cancer are mostly based on textual clinical information such as staging, age, histology, etc. Unlike existing methods that predicting on the single modality, we observe that a human clinician usually takes multimodal data such as text clinical data and visual scans to estimate survival time.
Motivated by this, in this work, we contribute a smart cross-modality network for survival analysis network named Lite-ProSENet that simulates a human's manner of decision making. To be specific, Lite-ProSENet is a two-tower architecture that takes the clinical data and the CT scans as inputs to create the survival prediction. The textural tower is responsible for modelling the clinical data, we build a light transformer using multi-head self-attention as our textural tower. The visual tower (namely ProSENet) is responsible for extracting features from the CT scans. The backbone of ProSENet is a 3D Resnet that works together with several repeatable building block named 3D-SE Resblock for a compact feature extraction. Our 3D-SE Resblock is composed of a 3D channel  “Squeeze-and-Excitation” (SE) block and a temporal SE block. The purpose of 3D-SE Resblock is to adaptively select the valuable features from CT scans. 
Besides, to further filter out the redundant information among CT scans, we develop a simple but effective frame difference mechanism that takes the performance of our model to the new state- of-the-art. 
Extensive experiments were conducted using data from 422 NSCLC patients from The Cancer Imaging Archive (TCIA). The results show that our Lite-ProSENet outperforms favorably again all comparison methods and achieves the new state of the art with the 89.3\% on concordance. The code will be made publicly available. 
\end{abstract}

%Another visual tower uses the 3D Resnet as its backbone and is responsible for extracting features from the CT scans. To adaptively select the valuable features from CT scans, we develop a 3D channel  “Squeeze-and-Excitation” (SE) block and a temporal SE block, working together with the Resblock to form the 3D- SE Resblock. Taking the 3D- SE Resblock as the basic building block, we develop our visual tower called ProSENet. To further filter out the redundant information among CT scans, we develop a simple but effective frame difference mechanism that takes our performance to the new state- of-the-art. Finally, the Lite-Transformer and ProSENet together with a prediction head form our Lite-ProSENet.

%\begin{graphicalabstract}
%%\includegraphics{figs/grabs.pdf} %%图片摘要地址路径
%\end{graphicalabstract}

%%高亮
%\begin{highlights}
%\item highlights 1.
%\item highlights 2.
%\item highlights 3.
%\end{highlights}

%% 关键词
\begin{keywords}
Multimodal learning \sep 
NSCLC \sep 
Survival analysis \sep 
Transformer
\end{keywords}

% 此指令为生成标题格式，不可删除
\maketitle

%%%%%%%%%%%%%%%%%%%%%%%%%%%%%%%%%%%%%%%%%%%%%%%%%%%%%%%%%%%%%%%%%%%%%%%%%%%%%%%%
\section{INTRODUCTION}

Lung cancer is one of the most dangerous diseases, the overall five-year survival rate for lung cancer (LC) is even less than 20\%. Most lung cancers can be divided into two broad histological subtypes: non-small cell lung cancer (NS-

\noindent CLC) and small cell lung cancer (SCLC). Compared to SCLC, NSCLC accounts for the majority of diagnoses and is less aggressive. NSCLC spreads and grows more slowly than SCLC and causes few or no symptoms until it is advanced. As a result, patients are usually not detected until it is at a later stage. And it has caused millions of deaths in both women and men. Lung cancer survival analysis, or prognostication, of lung cancer attempts to model the time range for a given event of interest (biological death), from the beginning of follow-up until the occurrence of the event. The survival model is an estimate of how lung cancer will develop, and it can reveal the relationship between prognostic factors and the disease. Using the accurate prognostic models, doctors can determine the most likely development(s) of the patient's cancer. To improve predictive accuracy and automate the NSCLC survival analysis process, as well as to assist medical experts develop precise treatment solutions, we aim to explore a novel method for NSCLC survival analysis.

\begin{figure}
    \centering
    \includegraphics[width=1.0\columnwidth]{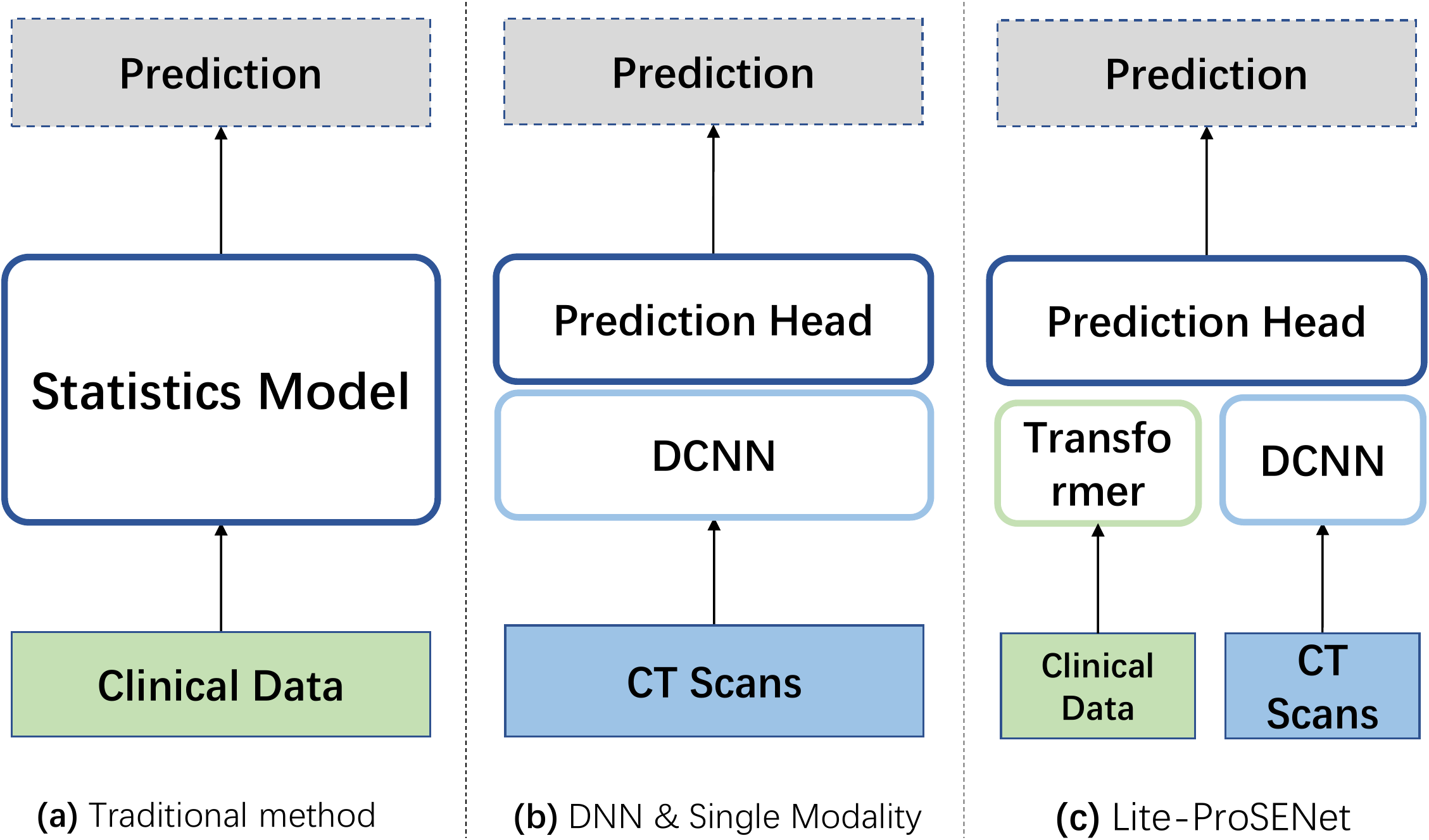}
    \caption{The architecture comparison of existing methods and our proposed Lite-ProSENet.}
    \label{first_fig}
\end{figure}

Traditional statistical methods for survival analysis leverage structured data from comprehensive medical examinations of a patient. Traditional methods mainly include time-to-event modelling tools such as the Cox proportional hazards (Cox- PH) model~\cite{cox1972regression}, the accelerated failure time (AFT) model~\cite{2020ACCELERATED}, the Kaplan-Meier~\cite{kaplan1958nonparametric}, etc. Besides, survival analysis based on machine learning is also a popular branch, such as survival trees~\cite{TM2004Bagging, 2013randomSurvivalForest}, Bayesian methods~\cite{2002Bayesian, 2011Bayesian} and Support Vector Regression~\cite{2008A, 2008Support} etc. These models assume that the hazard ratio between two subjects is constant over time~\cite{kvamme2019continuous} and estimate either a risk score or the time-to-event distribution. However, when implemented in clinical practise, the interaction between covariates is complex and all these methods focus only on structured data, overlooking the enormous information within the unstructured data such as CT scans. Moreover, medical experts usually have to dedicate extesive efforts in introducing the hand-crafted features. Recently, some works have been proposed using pathological images and demonstrating the effectiveness of features from images. However, pathological images require a lung biopsy. Moreover, the process of obtaining pathological images is invasive and associated with potential health risks such as pneumothorax, bleeding, infection, systemic air embolism and other side effects, which is a consequence of abnormal results of non-invasive computed tomography (CT) for lung cancer screening.

Artificial intelligence has witnessed rapid progress in the last decade. With the development of deep learning techniques, it has accomplished remarkable success in various fields of research such as natural language processing, computer vision, etc. As a cutting-edge technology, deep learning has the potential to offer great potential for medical diagnostics. Some of the most innovative and novel deep learning methods have been successfully applied to the diagnosis of lung cancer using CT images and the performance even outperformed human experts~\cite{esteva2017dermatologist} ~\cite{litjens2017survey} ~\cite{xing2017deep}~\cite{ILSVRC15}. However, existing deep learning-based methods only consider the structured or visual cues. In contrast, medical experts often consider clinical data and visual information such as CT images together to make a comprehensive decision. Consequently, the prediction of current methods is not reliable and credible enough. To address this weakness, we proposed a novel multimodal paradigm for lung cancer survival analysis inspired by the success of deep learning.

To successfully build such a multimodal network, the first challenge is to encode the task-friendly features from different modalities. The form of the clinical data looks like some discrete symbols in the diagram, but they are essentially highly correlated with each other in essence. Therefore, to find correlations between different words, a Light transformer network is proposed for processing the textual clinical data. The core building block of our model is the multi-head self-attention~\cite{DBLP:conf/nips/VaswaniSPUJGKP17}. Moreover, self-attention mechanism is capable of correlate different disease factors to capture more information. 

CT slices contain rich spatial and temporal information. In our previous work~\cite{DBLP:conf/smc/WuMHLS21}, we adopt 3D Resnet as a backbone for feature extraction. However, we found that there is too much redundant information in both the spatial and temporal dimensions, which severely prevents the model from perceiving the most important components in the visual information. To alleviate this problem, we for the first time propose a 3D channel SE block and a 3D temporal SE block. Both blocks are integrated into the original residual module to form an architecture specifically for NSCLC prognosis, namely ProgSE-Net. Moreover, we also observe that the pixels in adjacent CT slices are similar or the same in most cases. To support the ProgSE-Net, we further propose a mechanism of frame difference. The proposed frame difference creates two additional CT slices by subtracting the adjacent pixels in two directions, which is an effective strategy in our practise.

Finally, considering the above, we develop the first multimodal network for NSCLC survival analysis, which takes Deep Learning-based NSCLC survival analysis one step forward by simultaneously considering the textual clinical data and the visual CT clues. As shown in Fig.\ref{first_fig}, our network is a two-tower paradigm, \emph{i.e.}, clinical tower and a visual tower. The clinical tower is responsible for encoding the clinical data, while the visual tower aims to extract the visual representation from the CT images. Finally, the prediction head fuses the cross-modality features and provides the time prediction.

In summary, the key contributions associated with the Lite-ProSENet can be highlighted as follows: 
\begin{itemize}
	\item The first application of a two-tower DNN to survival analysis of NSCLC using structured data and CT images simultaneously. 
	\item The first application of transformer and 3DSE-Net block to multimodal clinical data for disease prognosis.
	\item Results on benchmark and real-world clinical datasets demonstrate that Lite-ProSENet outperforms SOTA methods by a substantial margin. %(12\% on concordance).
\end{itemize}

The rest of this paper is organized as follows: In section II, we present related work on survival analysis of NSCLC, including traditional methods and deep learning-based practice. In section III, we elaborate the details of our proposed Lite-ProSENet. In section IV, we present the experiments and ablation studies. We discuss some choices when building our network and the included hyper-parameters in section V. Finally, the conclusion and future work are given in section VI. In the following, the details of each section will be given.

\section{Related Works}
%In this section, we give an overview of the related work, and highlight the correlation to our contributions.
%\subsection{Related Works}
%refers to paper "Adversarial Time-to-Event Modeling" 
In this section, we give an overview of the traditional statistical methods and deep convolutional neural networks, then highlight the correlation to our contributions.

\subsection{Statistical Methods} 

Conventional statistical methods for NSCLC survival analysis only use the textual modality and involve modelling time to an event. They can be divided into three types: non-parametric, semi-parametric and parametric methods. Kaplan-Meier analysis (KM) ~\cite{2016An} is a typical non-parametric approach to survival outcomes. KM Analysis is suitable for small data sets with a more accurate analysis cannot include multiple variables. Life table ~\cite{1972Regression} is a simple statistical method that appropriate for large data sets and has been successfully applied to European lung cancer patients ~\cite{janssen1998variation}. The Nelson-Aalen estimator (NA) ~\cite{2013The} is a non-parametric estimator of the cumulative hazard function (CHF) for censored data. NA estimator directly estimates the hazard probability. As for semi-parametric method, the distribution of survival is not required. For example, the Cox regression model is used in ~\cite{port2003tumor}, which discovered the critical factor that has a greater impact on survival analysis in lung cancer. The Cox proportional hazards model~\cite{1972Regression} is the most commonly used model in survival analysis and the baseline hazard function is not specified. Coxboost can be applied to high-dimensional data to fit the sparse survival models. Better than the regular gradient boosting approach (RGBA), coxboost can update each step with a flexible set of candidate variables~\cite{2008Allowing}. The parametric method is easy to interpret and can provide a more efficient and accurate result when the distribution of survival time follows a certain distribution. But it leads to inconsistencies and can provide sub-optimal results if the distribution is violated. The Tobit model~\cite{1956Estimation}, for example, is one of the earliest attempts to extend linear regression with the Gaussian distribution for data analysis with censored observations. Buckley-James (BJ) regression ~\cite{1979Linear, 2008Doubly} uses least squares as an empirical loss function and can be applied to high-dimensional survival data. BJ regression is an accelerated failure time model. Bayesian survival analysis~\cite{2011Bayesian,2015Bayesian,2015Survival} encodes the assumption via prior distribution.
%25, 27, 28, 31, 33
\subsection{DNN based Methods} 
Image-based techniques for survival analysis of lung cancer normally adopt histopathological images. The work of ~\cite{DBLP:conf/miccai/YaoWZH16} was the first to use a deep learning approach to classify cell subtypes. The work found that survival models built from clinical imaging biomarkers had better predictive power than methods using molecular profiling data and traditional imaging biomarkers. Using machine learning methods, H. Wang et al, proposed a framework~\cite{2014Novel} and found a set of diagnostic image markers highly correlated with NSCL-C subtype classification. The work of Kun-Hsing Yu et al~\cite{DBLP:conf/amia/YuZBARRS17}, extracts 9,879 quantitative image features and uses regularised machine learning methods to distinguish short-term survivors from long-term survivors. In the work of Xinliang Zhu et al~\cite{DBLP:conf/bibm/ZhuYH16}, a deep convolutional neural network for survival analysis (DeepConvSurv) with pathological images was proposed for the first time. The deep layers in the proposed model could represent more abstract information, and hand-crafted features is not required from the images. The mentioned methods cannot learn discriminative patterns directly from Whole Slide Histopathological Images and some of them predict the survival status of patients using hand-crafted features extracted from manually labelled small discriminative patches. In the work of~\cite{2017WSISA}, an annotation-free method for survival prediction based on whole slide histopathology images was proposed for the first time.

In summary, traditional statistical methods tend to use textual data with limited information. In recent years, with the development of deep learning, more work has begun to explore methods that use histopathology images. However, it is invasive to obtain the images. There is a work that uses CT images but with hand-crafted features that require instructions from medical experts. Moreover, all these methods only use single modality and ignore the complementary information that comes from multimodality. Therefore, to capture the underlying complex relationships between multimodality medical testing results and NSCLC survival time, we proposed a non-invasive, fully automated DNN method to improve the prediction accuracy of NSCLC prognosis.

In this work, we propose a multimodal deep learning framework with 3D-ResNet for better individualised survival prediction of NSCLC, which can fully utilise and analyse information gathered from all types of data sources, such as CT images and clinical information.
%SEC3 methodology
%SYSTEM OVERVIEW -- brief intro about the structure
%SUBSECTION1 Light transformer -- embed, multihead self att, 
%SUBSECTION2 ProSE-Net 
%subsubsec channel se-block
%subsubsec temporal se-block
%subsubsec frame difference
%SUBSECTION3 multimodal feature fusion&prediction
%SUBSECTION3 network optimization -- loss mse,L2 penalty,date normalization
%figure1 system overview 
%figure 2 temporal se 
%figure 3 channel se
%SEC4 experiment
%..
%..
%..
%ablation study
\section{Methodology}
The proposed method is a two-tower architectural model. In this section, we describe details within the model for NSCL-C prognosis. 
%self att, multihead att, 
\subsection{The Architecture of Lite-ProSENet}

\begin{figure*}[t] 
\centering 
\captionsetup{labelfont=bf}
		\includegraphics[width=1\linewidth]{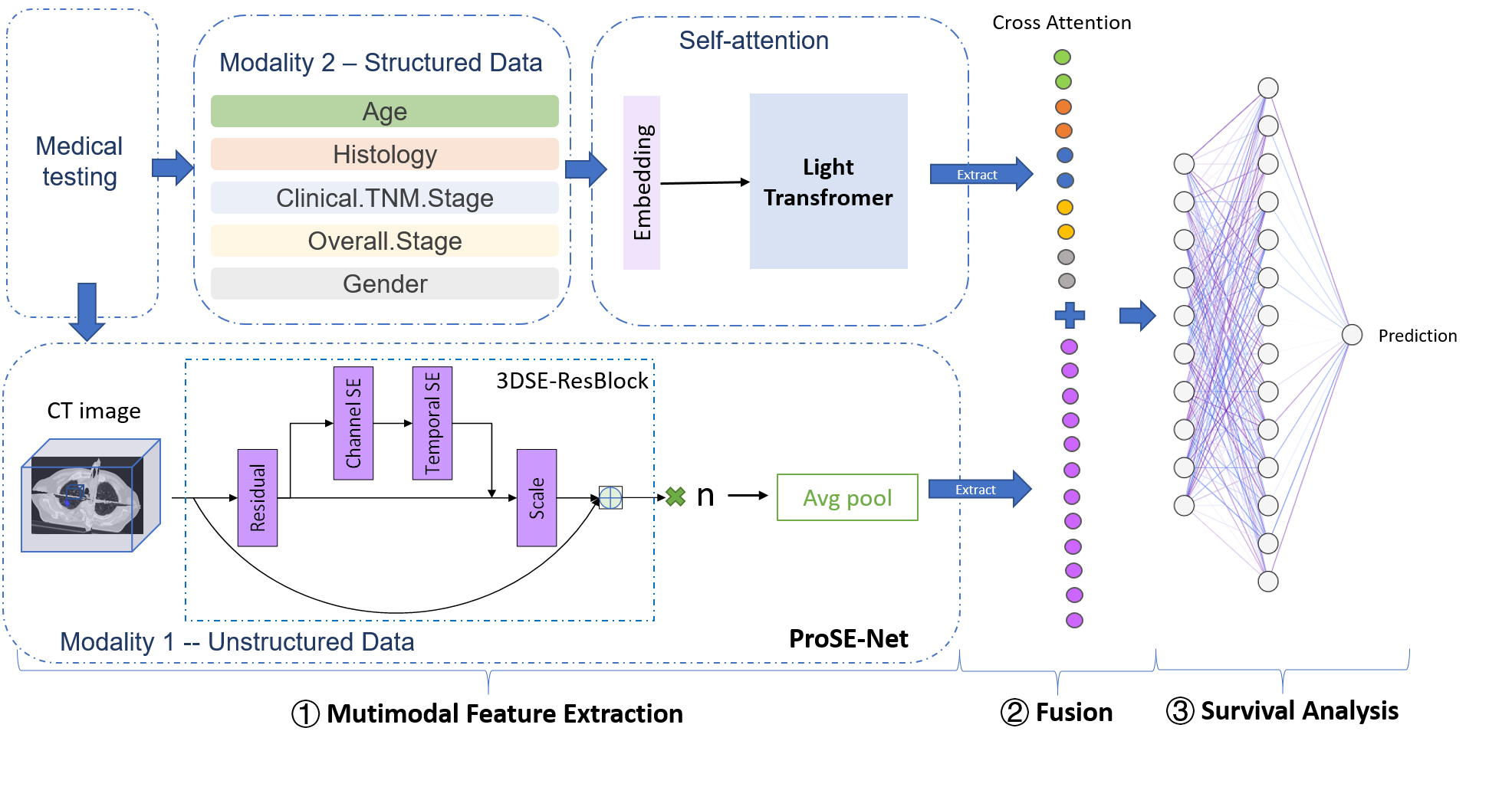}
		\vspace{-0.8cm}
\caption{A two-tower DNN model for learning similarity between textual clinical data and CT image representations.}
\label{tower_overview}
\end{figure*}

Clinical data and visual CT images both contain rich information but lie in different spaces, as a result, the information from different modalities cannot be integrated directly to give a comprehensive representation. To perform an effective feature fusion and alignment, we devise our model as a two-tower architecture, whose effectiveness has been well validated in the cross-modality learning field~\cite{2019Sampling, 2016Learning, 2016Smart, 2018Learning} . Figure~\ref{tower_overview} gives the overall illustration of our framework, the proposed Lite-ProSENet contains two towers, \emph{i.e.,} Lite-Transformer and ProSENet. Given a piece of data $d$, which is composed by the clinical data $c$, the CT images $I$ and survival time $t$, \emph{i.e.,} $d=\{(c,I),t\}$. The clinical data $c$ is first fed into an embedding layer to obtain the dense representation, and then pass through the 
light transformer to get the effective features.

CT images $I$ are first fed into the ProSENet for the feature extraction. The following prediction module fuses the features from different modalities and give the survival prediction $\hat{t}$ based on the multi-modality feature. Finally, the parameters of two towers are jointly optimized by minimizing the distance between the survival time prediction $\hat{t}$ and ground-truth one $t$.   In the following, we will 
illustrate the details of each component of our network.

\iffalse
DeepMMSA framework consists of three modules: 
{
	\begin{itemize}
		\item Multimodal feature (CT image features and Clinical record features) extraction.
		% \item Clinical record feature extraction network/model
		\item Multimodal feature fusion.
		\item Survival analysis.
	\end{itemize}
}
\fi
\subsection{Light Transformer}
%As  shown in the Fig. \ref{lite_trans}, the upper branch maps the input clinical sequence of symbol representation, $m=\{m_1, m_2, ..., m_i\}$, to a sequence of continuous representation, $n=\{n_1, n_2, ..., n_i\}$. Given n, the decoder then generates an output sequence, $o=\{o_1, o_2, ..., o_m\}$, of symbols one element at a time. At each step the model is auto-regressive~\cite{...}, consuming the previously generated symbols as additional input when generating the next.

As shown in Fig.~\ref{lite_trans}, the raw items in clinical data is first fed into an embedding layer to get a dense representation, then, the dense representations are fed into the multi-head self-attention layers to get the clinical features.

\vspace{0.2cm}
\noindent\textbf{Clinical Embedding.} A piece of clinical data usually contains several items, $c_i (i = 1, 2, ... m.)$, where $c_i$ is a kind of clinical item.  To better represent the raw clinical embedding, we assign each clinical item a dense feature using the popular embedding technique. We first give the initial item representation by the one-hot encoding, and then a matrix is introduced to project the initial representation to a dense feature:
\begin{equation}
    \label{first_eq}
    c_i = \text{one\_hot($c_i$)} \times W
\end{equation}
where one\_hot($\cdot$) is the function that project the item to a one hot vector, $W\in R^{V\times d}$ is the learnable map weight, $V$ is the item vocabulary size, and the $d$ is the dimension of dense representation. For the sake of symbol simplicity, we still use $c_i$ to denote the dense vector of item $c_i$.

\vspace{0.2cm}
\noindent\textbf{Multi-Head Self Attention.}
We adopt multi-head self attention in our model, which allows the model to jointly attend to information from different representation subspaces at different positions. Multi-head attention is an extension of self-attention, but repeat the attention mechanism several times. 

\begin{figure}[t]
	\centering
	\includegraphics[width=0.95\columnwidth]{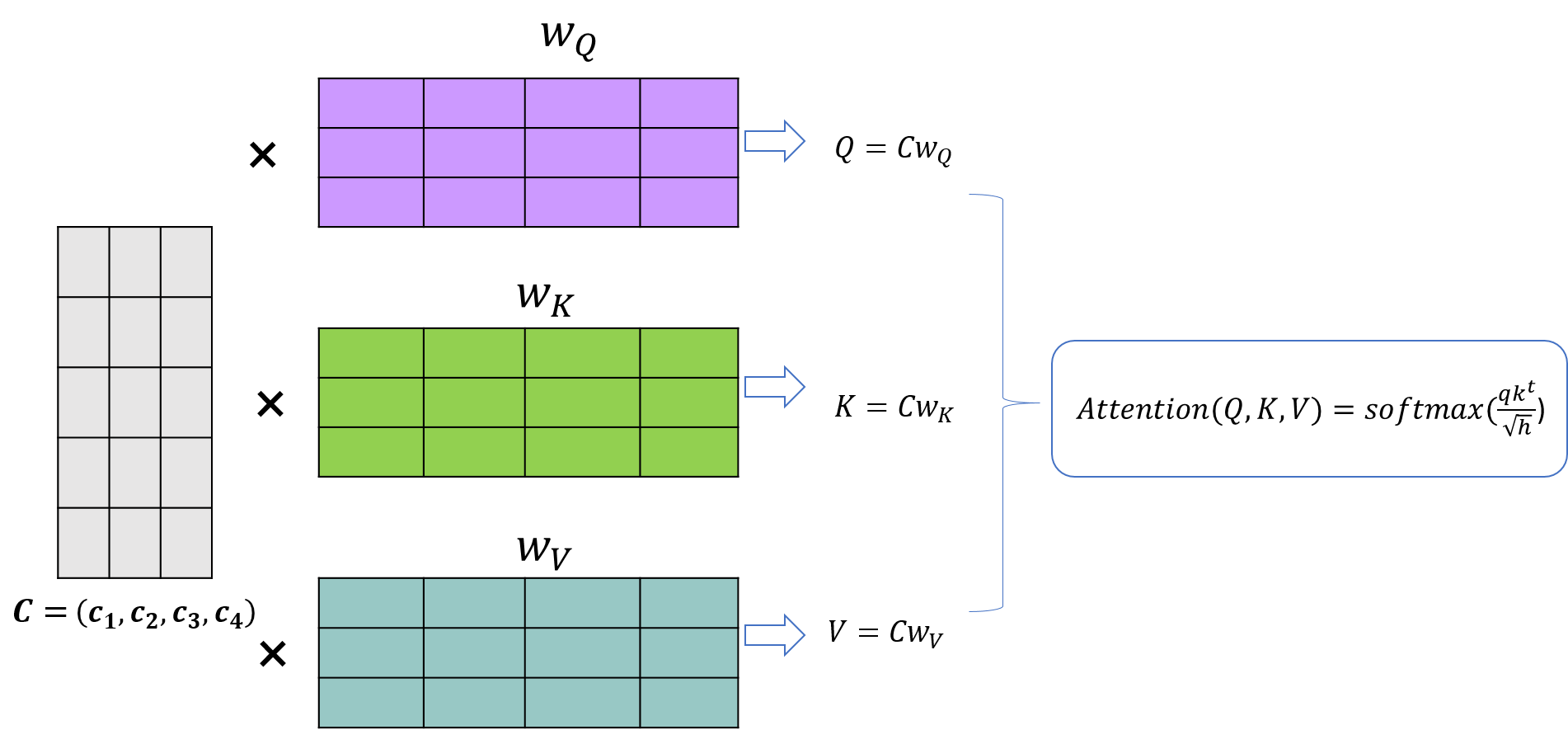}
	\caption{The process of applying self-attention layer to the Query, Value and Key matrices.}
	\label{self_att1}
\end{figure}

Each time, the transformer uses three different representations: the Queries, Keys and Values generate from the fully-connected layers. Fig.\ref{self_att1} illustrates the whole process of self-attention mechanism.  Let $C\in R^{m\times d}$ be the matrix formed by the item embeddings of clinical data $c$, mathematically, the outputs $S$ by the computation of self-attention can be expressed as:

\begin{figure}[t]
	\centering
	\includegraphics[width=0.95\columnwidth]{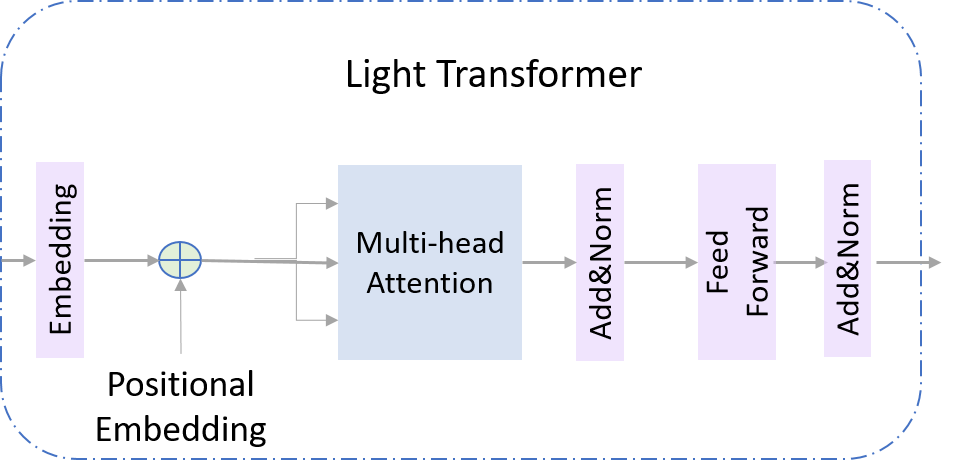}
	\caption{The Light Transformer - model architecture.}
	\label{lite_trans}
\end{figure}

\begin{equation}\label{selfattn}
\begin{split}
  &q = CW_q,\quad k = CW_k, \quad v=CW_v,\\
  &S = \text{softmax}(\frac{qk^T}{\sqrt{h}}),\quad\quad A = Sv.
\end{split}
\end{equation}
where $W_q, W_k, W_v\in R^{d\times h}$ are the learnable parameters, $h$ is the embedding dimension. Taking the self-attention (SA) as the basic block, the multi-head self-attention (MSA) is given by repeating the SA several times, and the outputs from different heads are concatenated together. Finally, the architecture of our lite-transformer is given as follows:
\begin{align}
A_{k} &= MSA(LN(A_{k-1})) + A_{k-1},  \qquad k =1,2,..K\\
A^{'}_{k} &= MLP(LN(A_{k})) + A_{k}, \qquad\quad\quad k=1,2,...K\\
T &= LN(A^{'}_K)
\end{align}
where $A_0=C$, $LN(\cdot)$ is the layer normalization, $T$ indicates the final clinical features, $K$ is the total MHA layers.

%3 different weight matrices are multiply with the input to get the Query, Value and Key matrices respectively and we can get Q, K, V matrices. There the resulted dimension will be smaller. After that, we applying a normalization layer and forming a residual skip connection. In Layer Normalization (LN), the mean and variance are computed across channels and spatial dims. The skip connections give our model the ability to allow the representations of different levels of processing to interact.

\subsection{ProSE-Net}
\label{visual_branch}
ProSE-Net is the model that learns unstructured data representation through a 3DResnet based network with several repeatable 3DSE-ResBlocks that is composed of a residual block, followed by a Channel SE-block and a Temporal SE-block. Such design can effectively improve the representational power of ProSENet. The key contributions of our ProSENet lie in the 3D channel SE block and temporal SE block, hereinafter, we will elaborate the details of these two modules.
%Specifically, the ProSE-Net is composed of a stack of N = xx identical 3DSE-ResBlocks. 

%Image features extracted by ProSE-Net and structure clinical data features extracted by from Light transformer are concate together using cross-attention for a better fusion. 

%CT images are informative, however, too much pixels are duplicated. To remedy this issue, we propose 3D SE block, which is composed of channel SE-block and temporal SE-block, to surpress the redundant information along channel and temporal dimensions.

\vspace{0.2cm}
\noindent\textbf{Channel SE-block.}
Channel SE-block targets to produce a compact feature via a squeeze-and-excitation operation along the channel dimension. Let $F\in R^{f\times c\times h\times w}$ be an arbitrary feature map, channel SE-block first performs ``Squeeze'' operation:
\begin{equation}
    p = \frac{1}{fhw}\sum_{i=1}^{f}\sum_{j=1}^{h}\sum_{k=1}^{w}F_{i,:,j,k}
\end{equation}
where $p\in R^{c}$.

Excitation operation first introduces two full-connected layers to perform a interaction between different channels, and a sigmoid is introduced to produce a information filter:
\begin{equation}\label{gate}
g=\text{sigmoid}(W_1\text{ReLU}(W_{2}p)),
\end{equation}
where $W_1 \in \mathbb{R}^{c \times \frac{c}{r}}, W_2 \in \mathbb{R}^{\frac{c}{r} \times c}$. $g\in R^{c}$ would server as the gate to perform information selection and the channel feature in $F$ would be updated as follow:
\begin{equation}\label{naiveSE}
    F^c = [F_{:,1,:,:}\times g_1, \cdots, F_{:,c,:,:}\times g_c]
\end{equation}

The common SE block only maintain the channel information and squeeze other dimensions, consequently, the important temporal information of 3D slices is also missed. To address this weakness, we augment the na\"ive SE block with a temporal excitation. First, the temporal dimension is remained when pooling the feature:
\begin{equation}
    p^t =  \frac{1}{hw}\sum_{j=1}^{h}\sum_{k=1}^{w}F_{j,k},
\end{equation}
$p^t\in R^{f\times c}$, we next produce a channel gate for each frame $g^t\in R^{f\times c}$, where the channel gate for $i-th$ frame is computed by Eq.~\ref{gate}:
\begin{equation}\label{gatet}
g^t_{i,:}=\text{sigmoid}(W_1\text{ReLU}(W_{2}p^t_{i,:})).
\end{equation}
we share the weights $W_1$ and $W_2$ when producing gates in each channel SE block. The goals for the weight sharing  stems from two aspects, the first is to propagate the information inside different views, building a lighter network with fewer paramters is the second reason. In our practice, sharing parameters can also promote the performance.

Finally, we fuse two types of gates and develop our full channel SE block as follows:
\begin{align}
         F^c &= [F_{1,1,:,:}\times G_{11}, F_{1,2,:,:}\times G_{12}, \cdots, F_{f,c,:,:}\times G_{fc}] \\
         G_{ij} &= g^t_{ij}\times g_{j}\label{fusegate}
\end{align}
where $G$ is the final gate filter, which is a combination of local view $g^t$ and global $g$ to perform a more reliable information filtering.

\begin{figure}[t]
\begin{center}
   \subfigure[Channel Squeeze-and-Excitation block]{
    \begin{minipage}[t]{0.86\linewidth}
    \label{channel_se}
        \centering
        \includegraphics[width=3.12in]{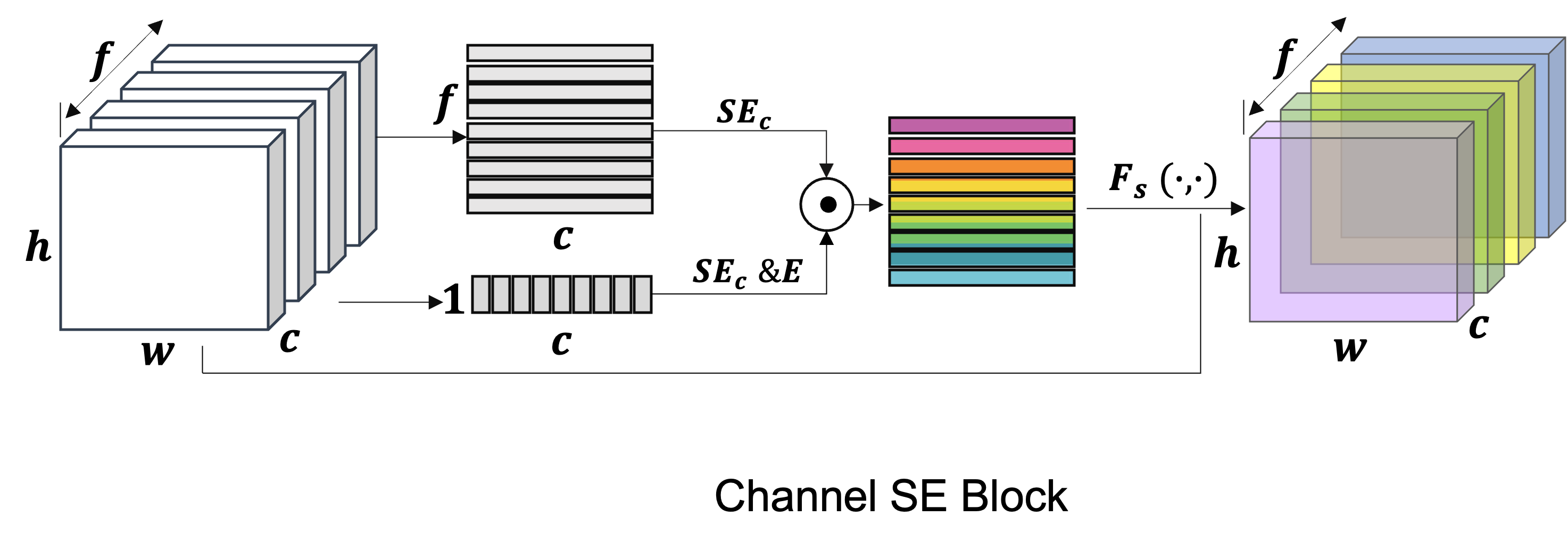}\\
    \end{minipage}
    \label{fig:fig3_a}
    }
    \vspace{-0.2cm}
    \subfigure[Temporal Squeeze-and-Excitation block]{
    \begin{minipage}[t]{0.86\linewidth}
    \label{temp_se}
        \centering
        \includegraphics[width=3.12in]{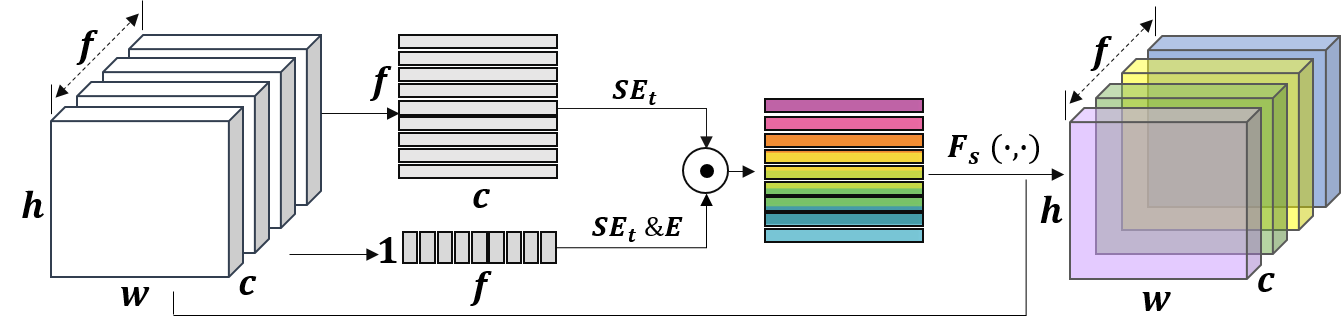}\\
    \end{minipage}
    \label{fig:fig3_b}
    }
\end{center}
\vspace{-0.2cm}
   \caption{Two types of Squeeze-and-Excitation blocks in our ProSENet.}
\label{temp_channel_se}
\end{figure}

\vspace{0.2cm}
\noindent\textbf{Temporal SE-block}
Similar with Channel SE-block, temporal SE-block also targets to filter out the redundant information but focuses on the temporal dimension. As shown in Figure \ref{temp_channel_se} (b), the computation procedure is analogous with the channel SE-block. The pooling is down along the channel-spatial dimension and spatial dimension, which provide the global and local frame information, respectively. Then, two types of gates are similarly produced following Eq.~\ref{gatet} and Eq.~\ref{gate}, which are then fused via Eq.~\ref{fusegate} and give the joint gate $G^{'}$. Finally, the temporal SE-block is formulated as:
\begin{equation}\label{tempGate}
    F^t = [F^c_{1,1,:,:}\times G^{'}_{11}, F^c_{1,2,:,:}\times G^{'}_{12}, \cdots, F^t_{f,c,:,:}\times G^{'}_{fc}].
\end{equation}
In our 3D SE block, the channel SE and temporal SE are stacked to achieve the information filtering along the channel and temporal dimensions, which forms our entire 3D SE-Resblock with the well-known 3D Resblock. The CT slices are first fed into the ProSENet to extract multi-dimension features and then pass through a 3D global average pooling to get the final feature $F_I$.

%can also help to provide the network with access to better global information and recalibrate filter responses. Different from Channel SE-block, the Temporal SE-block squeezed each features into a single numeric value using average pooling. Then a simple gating mechanism with a sigmoid activation is adopted. Then a bottleneck with two fully-connected (FC) layers is formed, aims to limit model complexity and aid generalisation of the system. After that, a RELU and    

\subsection{Multimodal feature fusion and prediction}
Given the clinical features $T$ from lite transformer and CT image feature $F_I$ from ProSENet, the next task is to fuse the multi-modality features and give the prediction of survival time $\hat{T}$. Thanks to the powerful features from our Lite transformer and ProSENet, we simply concatenate the cross-modal features and predict the survival time using a MLP, and a encouraging performance can be harvested in our practice:
\begin{equation}
    \hat{t} = MLP(concat([T, F_I])),
\end{equation}
where the $MLP(\cdot)$ is a two-layer full-connected layers, $concate(\cdot,\cdot)$ performs concatenation for the input two vectors.

\vspace{0.2cm}
\noindent\textbf{Enhance Prediction via Frame Difference.} We observe that although the CT images contain rich information, there are so many duplicated pixels between the CT slices, hindering the ProSENet to perceive the key information among the CT slices. To remedy this issue, we propose a simple yet effective mechanism, \emph{i.e.,} frame difference. The proposed frame difference performs a subtraction between two consecutive slices, such that the duplicated pixel could be ignored in the resulted slice. Following this idea, we perform the frame difference along two directions: forward and backward, the produced CT images are marked as $I^f,I^b$, respectively. Given this, our visual information are contain three types, \emph{i.e.,} the raw data $I$, frame difference along forward and backward direction $I^f$ and $I^b$. We then feed each of the visual information and the clinical data into our Lite-ProSENet, consequently, three time prediction could be given. Finally, we integrate the three predictions to produce the final result:
\begin{equation}\label{final_pred}
\bar{t} = \omega \hat{t} + (1-\omega)\frac{\hat{t}_f + \hat{t}_b}{2}
\end{equation}
where $\hat{t}_f$ and $\hat{t}_b$ are the survival prediction from the $(I^f,c)$ and $(I^b,c)$, respectively, $\omega$ is the trade-off weight, and $\bar{t}$ is the final prediction of survival time. 

\subsection{Network optimization}
%loss mse,L2 penalty,date normalization
With the final prediction $\bar{t}$ and the ground-truth survival time $t$, the network parameters are learned by minimizing the distance between the prediction and the ground-truth:
\begin{equation}\label{loss_func}
    \mathcal{L} = \frac{1}{b}\sum_{i=1}^b (\bar{t}-t)^2 +\lambda ||\mathcal{W}||_2,
\end{equation}
where $b$ is the batch size during training, $||\cdot||_2$ is the $l_2$ normalization, the second penalty is the parameter normalization, which is introduced to avoid overfitting, $W$ is all of the network parameters, and $\lambda$ is the trade-off hyper-parameter.

\section{Experiments}
We conduct extensive experiments based on NSCLC patients from TCIA to validate the performance of our proposed method with several state-of-the-art methods in terms of the prediction accuracy for the survival time for each patient. Besides, we also evaluate the prediction result by concordance. Afterward, we perform several ablation experiments regarding different network structures to determine the best structure. 

\subsection{Dataset}

In this work, we considered 422 NSCLC patients from TCIA to assess the proposed framework. For these patients pretreatment CT scans, manual delineation by a radiation oncologist of the 3D volume of the gross tumor volume and clinical outcome data are available~\cite{clark2013cancer}. The corresponding clinical data are also available in the same collection. The patients who had neither survival time nor event status were excluded from this work. 

\subsection{Data Preprocessing}
% \subsubsection{Data augmentation}
For CT images, we resize the raw data which is the 3D volume of the primary gross tumor volume into $96*96*8$. After that, we transform the range linearity into [0,1]. Then, to prevent overfitting, we perform data argumentation which includes three methods: rotate, swap, and flip. Then we get $422*8=3376$ samples, among which there are $373*8=2984$ uncensored samples and $49*8=392$ censored samples. 

Clinical data contains categorical data and non-categorical data. Firstly, missing value is a common problem in medical data and may pose difficulties for data analyzing and modeling. Specifically, in our dataset, the 'age' category contains a few missing values. After observing the data, we find that the age of patients in the dataset is close to each other. Thus, we impute the mean value and fill it into the missing value. Afterward, in order to fit into our model, we use the one-hot encoder to encode categorical data into numbers, which allows the representation of categorical data to be more expressive. 

Then, we use the min-max feature scaling method and standard score method to perform data normalization, such as age and survival time. For input $x$, the min-max feature scaling method's output is:
\begin{equation}
x'=\frac{x-x_{min}}{x_{max}-x_{min}}
\end{equation}
and the standard score method's output is:
\begin{equation}
x'=\frac{x-mean(x)}{std(x)}
\end{equation}
where $std$ is the standard deviation.
% where $\mu$ is the mean of $x$, and $\sigma$ is the standard deviation of $x$.

For a single patient with multiple tumors, we select the primary gross tumor volume ('GTV-1') to be processed in our work, while other tumors such as secondary tumor volumes denoted as 'GTV2', 'GTV3' to name just a few, which were occasionally present, were not considered in our work.

\subsection{Experiment Setup}
We train and evaluate the framework on the NSCLC-Radiomic dataset following 5-fold cross-validation with the patient-level split. We divided the dataset into training, validation, and testing data into 6:2:2 respectively. 
In Lite-transformer, the number of head in MHA is set as 3, and the total layers $K$ is configured in 5, more layers and heads bring limited performance gain but large parameters in our practice. In ProSENet, the ratio of channel and temporal SE are both set as 2, \emph{i.e.,} $r=2$. 
For hyperparameters tuning such as the penalty coefficient, we use the validation dataset to fine-tune and get the optimized hyperparameters. We config the $\omega$ in Eq.~\ref{final_pred} and the $\lambda$ in Eq.~\ref{loss_func} as 0.4 and 0.001 to trade-off different terms.  In the training process, we use 800 epochs in total with Adam as the optimizer. The batch size parameter is set as 64. The initial learning rate is set as 0.001, then the learning rate is decayed by 0.5 in every 40 epochs.
% ???The initial learning rate is 0.01, 0.001 after half the total number of epoch, and 0.0001 after epoch 120.

Since we use survival time as the label, not cumulative hazard. In the training and validation process, we only use the uncensored data for precise survival time and objective function calculation, and in the testing process, we use all data for concordance evaluation and uncensored data for MAE evaluation. 

Since this is the first work to use a multimodal framework for NSCLC survival analysis, we implement several state of the art survival analysis methods as baselines to compare with our work. The baseline methods include Cox-time~\cite{kvamme2019time}), DeepHit~\cite{lee2018deephit}, CoxCC~\cite{kvamme2019time}, PC-Hazard~\cite{kvamme2019continuous} and the regular cox regression.

\setlength{\tabcolsep}{5.5mm}{
\begin{table*}[t]
 \caption{C-index and MAE comparison between Lite-ProSENet and comparison methods.}
\begin{tabular}{c|c|c|cc|cc}
\hline
\multirow{2}{*}{Methods} & \multirow{2}{*}{Invasive} & \multirow{2}{*}{DNN-based} & \multicolumn{2}{c|}{Modality}                & \multicolumn{2}{c}{Performance}             \\ \cline{4-7} 
                         &                           &                            & \multicolumn{1}{c|}{Textual}    & Visual     & \multicolumn{1}{c|}{C-index${\color{red}\uparrow}$}        & MAE${\color{red}\downarrow}$    \\ \hline
Cox-time~\cite{kvamme2019time}                 & -                         & -                          & \multicolumn{1}{c|}{\checkmark}          & -          & \multicolumn{1}{c|}{0.6152}         & 0.183  \\ %\hline
Cox-regression~\cite{1972Regression}           & -                         & -                          & \multicolumn{1}{c|}{\checkmark}          & -          & \multicolumn{1}{c|}{0.6009}         & 0.204  \\ %\hline
CoxCC~\cite{kvamme2019time}                      & -                         & -                          & \multicolumn{1}{c|}{\checkmark}          & -          & \multicolumn{1}{c|}{0.6120}         & 0.183  \\ %\hline
PC-Hazard~\cite{kvamme2019continuous}                & -                         & -                          & \multicolumn{1}{c|}{\checkmark}          & -          & \multicolumn{1}{c|}{0.191}          & 0.6094 \\ %\hline
DeepHit~\cite{lee2018deephit}                  & -                         & -                          & \multicolumn{1}{c|}{\checkmark}          & -          & \multicolumn{1}{c|}{0.6133}         & 0.183  \\ %\hline
DeepMMSA~\cite{DBLP:conf/smc/WuMHLS21}                 & -                         & \checkmark                          & \multicolumn{1}{c|}{\checkmark}          & \textbf{\checkmark} & \multicolumn{1}{c|}{0.6580}         & 0.162  \\ %\hline
LASSO-Cox~\cite{Tibshirani97thelasso}                & -                         & -                          & \multicolumn{1}{c|}{\checkmark}          & -          & \multicolumn{1}{c|}{0.6698}          & NA     \\ %\hline
Cox + SuperPC~\cite{bair2006prediction}         & \checkmark                         & -                          & \multicolumn{1}{c|}{-}          & \checkmark          & \multicolumn{1}{c|}{0.556}          & NA     \\ %\hline
Log-logistic~\cite{2003Statistical}             & -                         & -                          & \multicolumn{1}{c|}{\checkmark}          & -          & \multicolumn{1}{c|}{0.5924}         & NA     \\ %\hline
BJ-EN~\cite{2010Buckley}                    & -                         & -                          & \multicolumn{1}{c|}{\checkmark}          & \textbf{-} & \multicolumn{1}{c|}{0.6646}         & NA     \\ %\hline
RSF~\cite{DBLP:journals/sadm/IshwaranKCM11}                      & -                         & -                          & \multicolumn{1}{c|}{\checkmark}          & -          & \multicolumn{1}{c|}{0.595}          & NA     \\ %\hline
MTLSA.V2~\cite{DBLP:conf/kdd/LiWYR16}                 & -                         & \textbf{-}                 & \multicolumn{1}{c|}{\checkmark}          & -          & \multicolumn{1}{c|}{0.680}          & NA     \\ %\hline
BoostCI~\cite{Mayr_2014}                  & -                         & \textbf{-}                 & \multicolumn{1}{c|}{\checkmark}         & \textbf{-} & \multicolumn{1}{c|}{0.6497}          & NA     \\ %\hline
WSISA~\cite{2017WSISA}                    & \checkmark                         & \checkmark                          & \multicolumn{1}{c|}{\textbf{-}} & \checkmark          & \multicolumn{1}{c|}{0.703} & NA     \\ %\hline
DeepSurv~\cite{DBLP:journals/corr/KatzmanSCBJK16}                 & -                         & \textbf{-}                 & \multicolumn{1}{c|}{\checkmark}          & \textbf{-} & \multicolumn{1}{c|}{0.602}          & NA     \\ %\hline
DeepConvSurv~\cite{zhu2016deep}            & \checkmark                         & \checkmark                          & \multicolumn{1}{c|}{-}          & \checkmark          & \multicolumn{1}{c|}{0.629}          & NA     \\ \hline
Lite-ProSENet             &                        & \checkmark                          & \checkmark         & \checkmark          & \multicolumn{1}{c|}{0.893}          & 0.043     \\ \hline
\end{tabular}
\label{main_res}
\end{table*}
}

\subsection{Quantitative Results}
In this subsection, we make a thorough comparison with both traditional and recent deep learning-based methods. The quantitative results of C-index and MAE are compared in Table~\ref{main_res}.

As shown in Table~\ref{main_res}, all the comparison methods except our previous work DeepMMSA only use the clinical data or the CT slices for prediction, for example, by building a survival function, Cox- regression can provide the probability that a certain event (e.g. death) occurs at a certain time $t$, the C-index of Cox- regression is only 0.601. In contrast, many experiments based on Deep Learning only use the visual information from CT scans. Although deep convolutional neural networks (DCNNs) are very powerful in feature extraction, the visual information alone is not reliable enough to accurately predict survival time. For example, the best C-index of deep learning-based methods only use visual CT is 0.703~\cite{2017WSISA}. Our previous work, DeepMMSA~\cite{DBLP:conf/smc/WuMHLS21} makes the first attempts to fuse the multimodal data using a two-tower framework. Although we found that multimodal inputs could boost the performance, the final results do not surpass the deep learning based methods using only visual information such as WSISA~\cite{2017WSISA}. This observation indicates that the straightforward network cannot work well for multimodal fusion. Consequently, we developed our Lite-ProSENet to build an effective multimodal network for survival analysis. Our Lite-ProSENet was able to achieve a C-index of 0.893, outperforming all comparative methods , which well validate the superiority of our method.

\subsection{Ablation Study}
To build an effective cross-modal survival model, we design our Lite transformer for clinical data and propose the 3D- SE Resblock to effectively model the visual CT slices. Furthermore, we propose a frame difference mechanism to promote our performance to the new state-of-the-art. In this subsection, we will verify the effectiveness of the above modules to support our claims through extensive experiments.

The results are reported in Table~\ref{abla}, where we systematically examine the contribution of each component, including the Lite-Transformer, the 3D- SE Resblock in ProSENet, and the mechanism of frame difference. In the baseline method ( no modules are equipped ), the Lite-Transformer is replaced by several MLP layers to form a similar parameters. As is shown in Table~\ref{abla}, the C-index of the baseline method is only 0.796, and the C-index improves when each module is equipped. For example, the baseline with Lite-Transformer could achieve a C-index of 0.824, and the 3D- SE Resblock helps the baseline to improve the C-index from 0.796 to 0.841. Applying any two modules simultaneously could improve the performance even further. If we apply 3D- SE Resblock and frame difference, we could attain the C-index of 0.873, which is a significant improvement. When all of three modules are configured, we harvest the best performance, whose C-index could reach the new state- of-the-art 0.893. The observation on MAE shows the consistent tendency.

As one of our main motivations for the Lite-ProSENet design, verifying the effectiveness of multi-modality modeling is also a critical aspect. We also investigate the benefits of multi-modality learning from this aspect. The results are also reported in Table~\ref{abla}, where $\text{Lite-ProSENet}_{V}$ and $\text{Lite-Pr-}$

\noindent $\text{oSENet}_{T}$ refer to the $\text{Lite-ProSENet}$ with visual tower and textural tower, respectively. We can observe that the network with any tower alone could not achieve satisfactory performance, the visual tower only achieves a C-index of 0.712. Although the 3D- SE block boosts the performance to 0.739, it is still not satisfactory. The observations of $\text{Lite-ProSENet}_{T}$ are also conclusive. The model with multi-modality learning could achieve a C-index of 0.796, which well demonstrates the importance of fusing the clinical data and the visual CT images for the survival time analysis.

\setlength{\tabcolsep}{3.0mm}{
\begin{table*}[]
    \centering
     \caption{Discuss the effectiveness of slices from frame difference, `$\checkmark$' and `-' means applying and not applying the corresponding modules.}
    \begin{tabular}{c|ccc|cc}
    \hline      
      &  Lite-Transformer  &  3D-SE Resblock & Frame Difference & C-index${\color{red}\uparrow}$ & MAE${\color{red}\downarrow}$\\
      \hline
     \multirow{8}*{Lite-ProSENet} & - & -  & -  & 0.796  & 0.121\\
         &  \checkmark &-  & - &0.824  & 0.108\\
         & - & \checkmark &-  &0.841   & 0.092\\
         & - & - &\checkmark & 0.837 & 0.103 \\
         \cline{2-6}
         &\checkmark &\checkmark  & - &0.859  & 0.086 \\
         & -& \checkmark&\checkmark &0.873 & 0.063\\
         &\checkmark &- &\checkmark &0.862 & 0.071\\
         \cline{2-6}
         &\checkmark &\checkmark &\checkmark &0.893 &0.043\\
     \hline
     \multirow{2}*{$\text{Lite-ProSENet}_{V}$} & - & - & - &0.712 & 0.223  \\
      & -&\checkmark &- &0.743 &0.187\\
     \hline
      \multirow{2}*{$\text{Lite-ProSENet}_{T}$} &- &- &- &0.739 &0.181  \\
      &\checkmark &- &- &0.761 & 0.149\\
     \hline
    \end{tabular}
   
    \label{abla}
\end{table*}
}

\section{Discussion}
In this section, we would give several discussion about the many choices when building our network, including the effect of the joint gate in our 3D SEResblock, the order of two SE blocks, the impact of the bi-directional frame difference. Besides the choices of several mechanisms, the hyper-parameters, $\omega$ in Eq.~\ref{final_pred} and $\lambda$ in Eq.~\ref{loss_func}, are also presented in this section.

\setlength{\tabcolsep}{1.0mm}{
\begin{table}[]
    \centering
     \caption{Discuss the importance of the global and local SE module in our ProSENet, `$\checkmark$' and `-' means applying and not applying the corresponding modules.}
    \begin{tabular}{c|c|cc|cc}
    \hline
      &  SE blocks & global SE  &  local SE & C-index${\color{red}\uparrow}$ & MAE${\color{red}\downarrow}$\\
      \hline
    \multirow{8}*{baseline} & \multirow{4}*{Channel SE} & - & -  & 0.826  & 0.058\\
         &  &\checkmark &-  & 0.842 &0.051\\
         &  &- & \checkmark & 0.867 & 0.047 \\
         & &\checkmark &\checkmark  &0.893 & 0.043\\
         \cline{2-6}
       & \multirow{4}*{Temporal SE} & - & -  & 0.819  & 0.061\\
     &     &  \checkmark &-  & 0.839 &0.053\\
      &   & - & \checkmark & 0.862 & 0.049 \\
       &  &\checkmark &\checkmark  &0.893 & 0.043\\
     \hline
    \end{tabular}
   
    \label{joint_gate}
\end{table}
}

\subsection{Validate the joint gate in 3D SEResblock.}
In the 3D SEReslock, we augment the channel SE and the temporal SE with the joint gate to perform the information filtering, more details can be found in subsection~\ref{visual_branch}. In this subsection, we would validate the effectiveness of our proposed joint gate.

The results are reported in Table~\ref{joint_gate}, we set the baseline as the network where visual tower is the na\"ive 3D Resnet, `global SE` refers to the gate is only built by the na\"ive SE block, for channel SE, the output of global SE is produced by the Eq.~\ref{naiveSE}\footnote{\noindent When studying the channel SE (temporal SE), we equip the full temporal SE block (channel SE).}. `local SE' indicates the gate is only build by the channel-wise or frame-wise information, for channel SE, the output of global SE is produced by replacing the $g$ in Eq.~\ref{naiveSE} with $g^t$ defined in Eq.~\ref{gatet}. Joint gate uses both the global and local SE block, \emph{i.e.,} our 3D SEResblock. As we can observed from Table~\ref{joint_gate}, SE block is an effective block, the system is benefit from both types of SE block. For channel SE block, when equipping the global SE block, the C-index is improved from 0.826 to 0.842, and the local SE block also boosts the perform from 0.826  to 0.867. When the joint gate is applied, the performance gets a significantly improvement, from 0.0.826 to 0.893. The observation of temporal SE block is also conclusive, which well validates the effectiveness of 3D-SE Resblock.

\begin{table}[]
    \centering
     \caption{The performance comparison of different stacking order of channel SE and temporal SE, where I and II mean ranking in the first and the second place, the meaning of $\checkmark$ and - is the same as Tabel~\ref{joint_gate}}
    \begin{tabular}{c|cc|cc}
    \hline
      &  Channel SE  &  Temporal SE & C-index${\color{red}\uparrow}$ & MAE${\color{red}\downarrow}$\\
      \hline
     \multirow{4}*{Lite-ProSENet}   & $\checkmark$ & - & 0.871 & 0.058\\
      & - & $\checkmark$ &0.879 &0.055 \\
    &  II & I  & 0.881 &0.049\\
     & I & II  & 0.893  & 0.043\\
      
     \hline
    \end{tabular}
   
    \label{order_discussion}
\end{table}

\subsection{Study the stacking order of two SE blocks.}
In our 3D- SE Resblock, the channel SE is applied first, and the temporal acts on the output of the channel SE block, as shown in Eq.~\ref{fusegate} and Eq.~\ref{tempGate}. In this subsection, we study the difference in performance between two SE blocks in different stacking order.

The performance comparison is given in Table~\ref{order_discussion}, we study two types of stacking order, \emph{i.e.,} channel SE first and then temporal SE second, and vice versa. As shown in Table~\ref{order_discussion}, the strategy of the channel SE first and the temporal SE second performs better. The C-index of channel SE first could reach 0.893, while the temporal SE first is worse, whose c-index is 0.881. Consequently, we first apply the channel SE in our network to achieve a better C-index. In addition to stacking order, in this subsection, we also investigate the importance of two SE blocks. As shown in Table~\ref{order_discussion}, using the channel SE or the temporal SE performs alone performs worse than using two SE blocks simultaneously with arbitrary stacking order, which validating the effectiveness of our channel and the temporal SE block.

\begin{table}[]
    \centering
     \caption{Discuss the effectiveness of slices from frame difference, `$\checkmark$' and `-' means applying and not applying the corresponding modules.}
    \begin{tabular}{c|cc|cc}
    \hline
      &  forward  &  backward & C-index${\color{red}\uparrow}$ & MAE${\color{red}\downarrow}$\\
      \hline
     \multirow{4}*{Lite-ProSENet} & - & -  & 0.854  & 0.058\\
         &  \checkmark &-  & 0.881 &0.046\\
         & - & \checkmark & 0.879 & 0.045 \\
         &\checkmark &\checkmark  &0.893 & 0.043\\
     \hline
    \end{tabular}
   
    \label{bidirectionFD}
\end{table}

\subsection{The effectiveness of the bi-directional frame difference.}
When predicting the final survival time, we introduce frame difference to filter out the redundant information between different CT slices. To further boost the performance, we perform bidirectional frame difference among CT images. In this subsection, we discuss the effectiveness of our bidirectional frame difference.

To thoroughly validate the effectiveness of the proposed frame difference, we study three cases, \emph{i.e.,}, only the frame difference along forward direction and backward direction, and the bi-directional frame difference. The results can be found in Table~\ref{bidirectionFD}, where the 'forward' and 'backward' mean the normal direction and the reverse direction, respectively. From Table~\ref{bidirectionFD} we can observe that both the'forward' and 'backward' frame difference can promote the performance.  When introducing the forward frame difference, the C-index gets improved from 0.854 to 0.881, and the backward frame difference can boost the C-index from 0.854 to 0.879. When we integrate the frame difference simultaneously in the forward and backward directions, we get the best C-index of 0.893. These observations well reveal that our proposed frame difference is an effective mechanism.

\subsection{Discussion about hyper-parameter $\omega$}
We introduce a trade-off parameter when integrating the prediction of normal CT slices and the bi-directional frame difference, as shown in Eq.~\ref{final_pred}. The $\omega$ is set as 0.4 by default. In this subsection, we would study the performance change when the hyper-parameter $\omega$ varied.

\begin{figure}[t]
\begin{center}
   \subfigure[The performance comparison when $\omega$ varies from 0 to 1.]{
    \begin{minipage}[t]{0.86\linewidth}
    \label{omega_dis}
        \centering
        \includegraphics[width=3.12in]{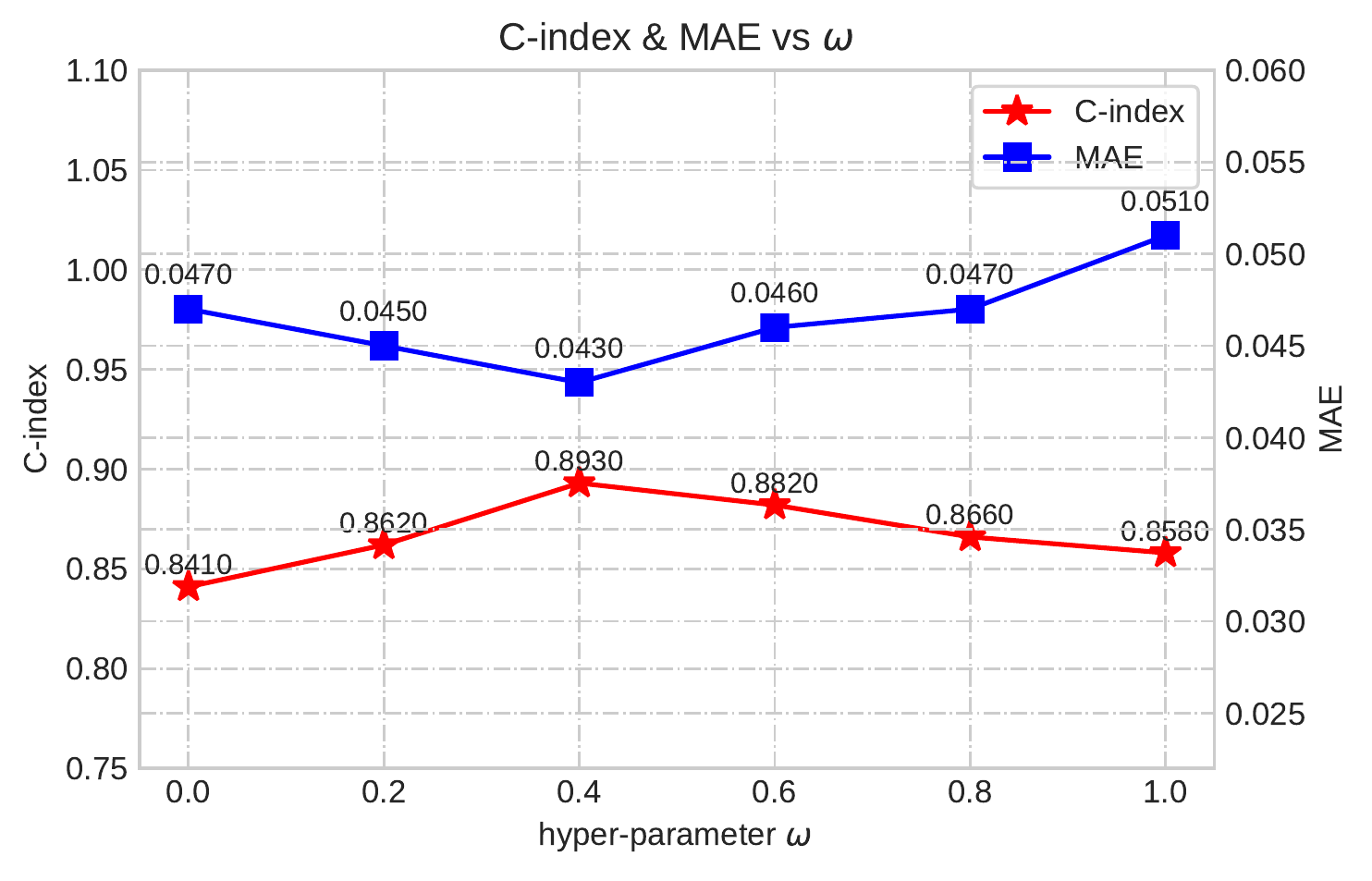}\\
    \end{minipage}
    }
    \vspace{-0.2cm}
    \subfigure[Performance comparison when $\lambda$ varies from 0 to 0.1.]{
    \begin{minipage}[t]{0.86\linewidth}
    \label{lambda_dis}
        \centering
        \includegraphics[width=3.12in]{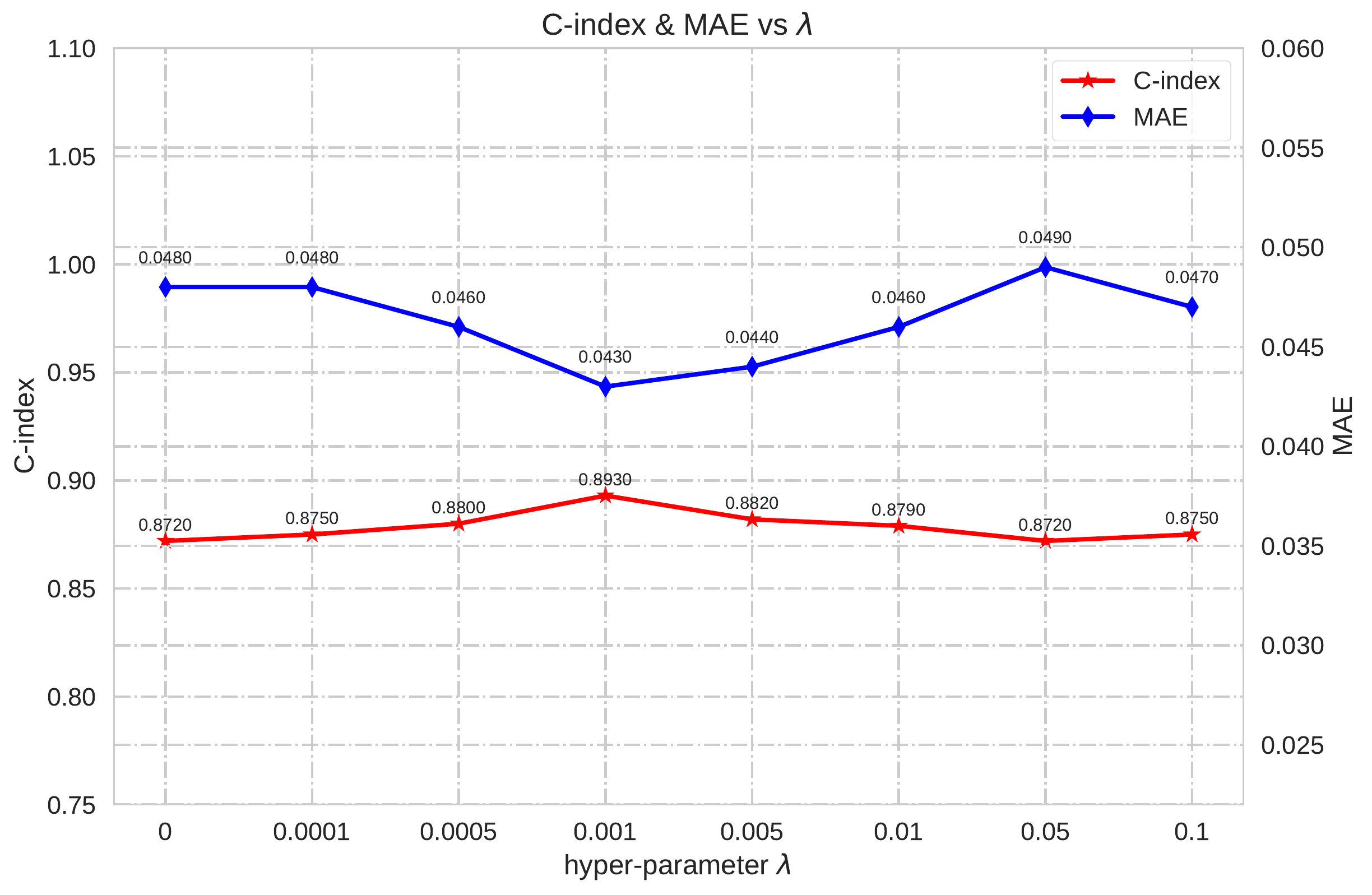}\\
    \end{minipage}
    }
\end{center}
\vspace{-0.2cm}
   \caption{Discussion about the hyper-parameters $\omega$ in Eq.~\ref{final_pred} and $\lambda$ in Eq.~\ref{loss_func}}
\end{figure}

Figure~\ref{omega_dis} shows the performance of our network when $\omega$ varied from 0 to 1 with step 0.2. We can observe from Figure~\ref{omega_dis} that the c-index and MSE loss show consistent trendency. When $\omega=0$, this means we only predict by the slices of frame difference and ignore the normal CT slice, this case does not achieve a satisfactory performance whose c-index is only 0.841, the reason for this observation may stem from too much information missed when completely abandoning the original CT slice. This guess is validated by the cases of $\omega\neq 0$. The case of $\omega=1$ means we does not apply the frame difference, such case also fail to achieve the best performance, revealing that the slices of frame difference are necessary for our network. The case of $\omega=0.4$ achieve the best performance, this means that slices from frame difference play an important role in the survival time prediction. Further enlarging the weight of frame difference does not promote the performance. Therefore, we fix $\omega$ as 0.4 in our network.

\subsection{Discussion about hyper-parameter $\lambda$.}
When training the network, we employ the popular parameter normalization strategy to avoid overfitting, \emph{i.e.,} the second term in Eq.~\ref{loss_func}, and introduce a hyperparameter $\lambda$ to balance the main loss and the parameter normalization. By default, we set $\lambda$ to 0.001. In this section, we will study the impact of the hyper-parameter $\lambda$ on the performance.

The changes of c-index and MSE loss are shown in Figure~\ref{lambda_dis}, where the y-axis represents performance and the x-axis represents $\lambda$. We study the performance under $\lambda=\{0, 0.0001,$
\noindent$0.0005,0.001,0.005,0.01,0.05,0.1\}$, from Figure~\ref{lambda_dis}, we can observe a clear trend. When $\lambda=0$, which means that we dispense with parameter normalization, the network does not achieve very good performance. Then, when we increase $\lambda$, the performance starts to increase. When $\lambda=0.001$, we were able to achieve the best c-index 0.893. If we increase $\lambda$ further, we cannot obtain more performance gains. In the case of $\lambda=0.4$, a good trade-off between the main loss and the parameter normalization is achieved.

\section{Conclusion and Future Work} 

This work contributes a powerful multimodal network for more accurate prediction of NSCLC survival, with the purpose of helping clinicians to develop timely treatment plans and improve patients' quality of life. Our method provides a new state-of-the-art result of 89.3\% on the C-index.
To well model the cross-modal data, we develop a two-tower network, with the textual tower responsible for the clinical data and the visual tower for the CT slices. Inspired by the success of the transformer in the NLP field, we propose a very light transformer using the core of self-attention. For the visual tower, we design a ProSENet based on the 3D- SE Resblock, where channel Squeeze-and-excitation and temporal Squeeze-and-excitation are proposed to suppress the redundant information among the CT slices. Besides, we further introduce a frame difference mechanism to help promote our network up to the new state-of-the-art in terms of C-index and MAE. In experiments, we conduct comparisons, ablation studies and discussions that well verify the superiority of our Lite-ProSENet. The practice of this work gives us much confidence about the deep learning-based survival analysis. We believe that the deep learning-based method has great potential to be realized for survival time analysis. In the future, we will further investigate this problem from the following two aspects:
\begin{itemize}
    \item \textbf{Effective fusion of cross-modal features.} In this work, the fusion of multimodal features is simple, we simply concatenate the features from Lite-transformer and ProSENet. In the future we will explore more effective fusion manners.
    
    \item \textbf{Borrow information from large-scale pretrained models.} Large-scale pretrained cross-modal models have shown great potential in many tasks, such as Visual question answering,  images captioning, cross-media retrieval, \emph{et al.} After training with millions of data, the large-scale models contain powerful knowledge, how to adapt these knowledge to survival time analysis is a promising direction. We will explore this direction in the future.
\end{itemize}

%%%%%%%%%%%%%%%%%%%%%%%%%%%%%%%%%%%%%%%%%%%%%%%%%%%%%%%%%%%%%%%%%%%%%%%%%%%%%%%%

%%%%%%%%%%%%%%%%%%%%%%%%%%%%%%%%%%%%%%%%%%%%%%%%%%%%%%%%%%%%%%%%%%%%%%%%%%%%%%%%

%%%%%%%%%%%%%%%%%%%%%%%%%%%%%%%%%%%%%%%%%%%%%%%%%%%%%%%%%%%%%%%%%%%%%%%%%%%%%%%%

%\begin{thebibliography}{99}

{\small
\bibliographystyle{ieeetr}
\bibliography{ijcai21}
}

%\end{thebibliography}

\end{document}